\documentclass[conference]{IEEEtran}
\IEEEoverridecommandlockouts

\usepackage{cite}
\usepackage{amsmath,amssymb,amsfonts}
\usepackage{algorithmic}
\usepackage{graphicx}
\usepackage{textcomp}
\usepackage{xcolor}
\usepackage{booktabs}

\begin{document}

\title{\Large \textbf{HYBRID MULTI-DIMENSIONAL MRI PROSTATE CANCER DETECTION VIA HADAMARD NETWORK BASED BIAS CORRECTION AND RESIDUAL NETWORKS}%
\thanks{This work was supported by the National Science Foundation (NSF) under Grant IDEAL 2217023, and by the National Institutes of Health (NIH) under Grants U01-CA268808, R42-CA244056, R01-CA227036, R41-CA244056, R01-CA17280, and 1S10OD018448-01. Additional support was provided by the Sanford J. Grossman Charitable Trust and the University of Chicago Medicine Comprehensive Cancer Center (Grant P30 CA014599-37).}
}

\author{
Emadeldeen Hamdan\textsuperscript{1},
Gorkem Durak\textsuperscript{2},
Muhammed Enes Tasci\textsuperscript{2},
Abel Lorente Campos\textsuperscript{3},
Aritrick Chatterjee\textsuperscript{3} \\
Roger Engelmann\textsuperscript{3},
Gregory Karczmar\textsuperscript{3},
Aytekin Oto\textsuperscript{3,*},
Ahmet Enis Cetin\textsuperscript{1,*},
Ulas Bagci\textsuperscript{2,*} \\ \\
\textsuperscript{1}Electrical and Computer Engineering Department, University of Illinois Chicago, USA \\
\textsuperscript{2}Machine and Hybrid Intelligence Lab, Northwestern University, USA \\
\textsuperscript{3}Department of Radiology, University of Chicago, USA \\
\textsuperscript{*}Co-Senior Authors
}

\maketitle
\begin{abstract}
Magnetic Resonance Imaging (MRI) is vital for prostate cancer (PCa) diagnosis. While advanced techniques such as Hybrid Multi-dimensional MRI (HM-MRI) have enhanced diagnostic capabilities, the significant need remains for robust, automated Artificial Intelligence (AI)-based detection methods. In this study, we combine quantitative HM-MRI of tissue composition with an AI-based neural network. We propose the Hadamard-Bias Network plus ResNet18 (HBR-Net-18), a two-stage AI framework for PCa detection. In the first stage, a Hadamard U-Net-based algorithm suppresses intensity inhomogeneities (bias fields) across six parametric HM-MRI maps generated via a Physics-Informed Autoencoder (PIA). In the second stage, a Residual Network (ResNet-18) performs patch-level classification. The framework utilizes overlapping 11-by-11 patches, incorporating both 2D intra-slice and 3D inter-slice (adjacent-slice) information to improve spatial consistency. Our experimental results demonstrate that HB-Net achieves balanced sensitivity and specificity, significantly outperforming conventional radiomics-based approaches and baseline CNN models, highlighting its potential for clinical deployment.

\end{abstract}
\section{Introduction}
\label{sec:intro}

Prostate cancer (PCa) remains the most common cancer among men worldwide and a leading cause of cancer-related mortality \cite{litwin2017diagnosis}. Magnetic Resonance Imaging (MRI) and Hybrid Multi-Dimensional MRI (HM-MRI) \cite{chatterjee2025hybrid} have emerged as central imaging modalities for PCa diagnosis due to their superior soft-tissue contrast, multi-parametric capabilities, and quantitative measures of tissue composition and characteristics. HM-MRI models and quantifies the unique MRI properties of the prostate tissue compartments—stroma, epithelium, and lumen—enabling noninvasive assessment of tissue composition and microstructure. 

Traditional approaches to prostate cancer detection and analysis include radiomics-based methods using various MRI modalities \cite{ferro2022radiomics,sun2019multiparametric}.
More recently, Artificial Intelligence (AI)–driven algorithms have become the foundation for rapid and automated cancer detection techniques \cite{saha2024artificial, giganti2025ai}. Convolutional Neural Networks (CNNs) have shown a strong ability to extract unique features from medical images, allowing the detection of subtle spatial patterns that may not be visually apparent to experts \cite{liu2024hybrid, hossain2025automated}.  However, their effectiveness is often limited by bias field inhomogeneities, signal variability across MRI sequences, and limited inter-slice generalization, which can result in inconsistent or unreliable cancer predictions.


To address these challenges, we formulate prostate cancer detection as a bias-aware, two-stage learning problem, where low-frequency intensity variations in quantitative MRI biomarkers are explicitly corrected prior to classification. We propose the Hadamard-Bias Network + ResNet-18 (HBR-Net-18), a modular AI framework operating on six tissue-based biomarkers generated from HM-MRI using a Physics-Informed Autoencoder (PIA)~\cite{gundogdu2025physics}.

\begin{figure*}[ht]
    \centering
    \includegraphics[width=\linewidth, height=7cm]{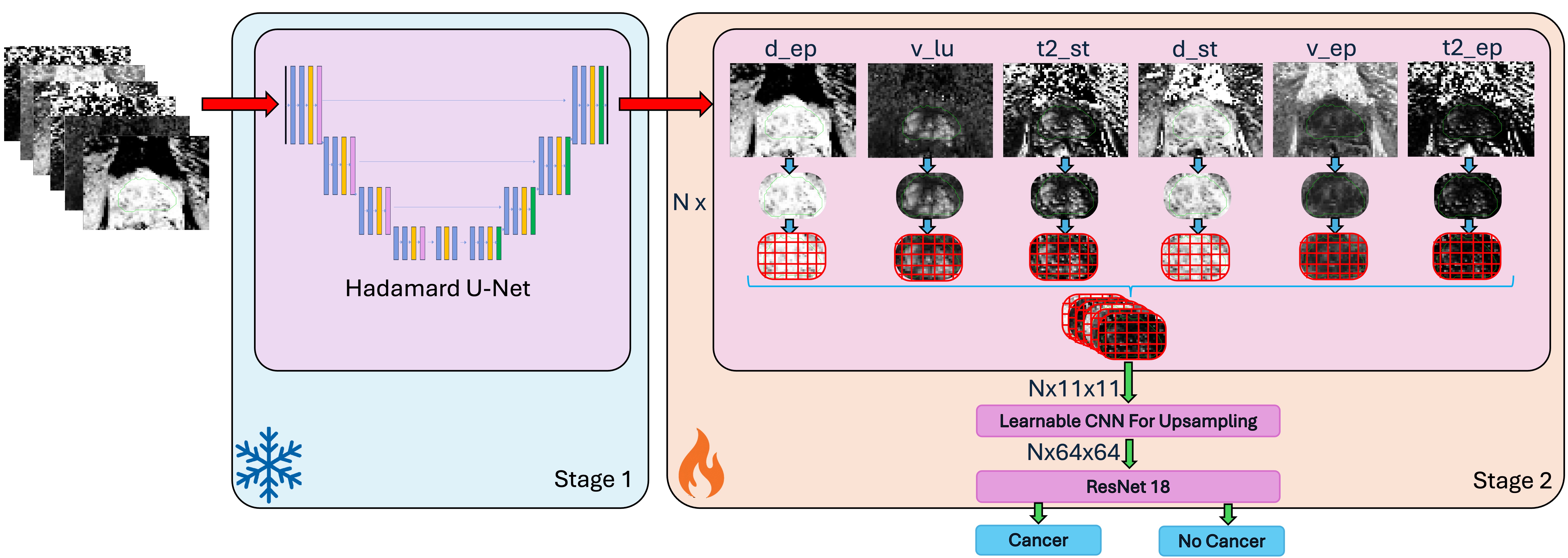}
    \caption{\textbf{Overview of the proposed HBR-Net-18 framework.}
    \textbf{Left}: Bias field correction stage based on a probabilistic Hadamard U-Net, designed to suppress low-frequency intensity inhomogeneities and noise across six quantitative tissue biomarkers generated by the Physics-Informed Autoencoder (PIA): epithelial volume fraction ($v_{ep}$), luminal water fraction ($v_{lu}$), epithelial diffusivity ($d_{ep}$), stromal diffusivity ($d_{st}$), epithelial T2 relaxation time ($t2_{ep}$), and stromal T2 relaxation time ($t2_{st}$).
    \textbf{Right}: Cancer detection stage, where corrected biomarker maps are confined to the prostate region, divided into overlapping $11\times11$ patches, and processed by a convolutional upsampling layer followed by a ResNet-18 classifier for patch-level prostate cancer detection.}
\label{fig:Model}
\end{figure*}

In the first stage, a learnable probabilistic Hadamard U-Net~\cite{zhu2024probabilistic} is trained to suppress noise and correct low-frequency intensity inhomogeneities within each biomarker map, while preserving diagnostically relevant spatial structures. Crucially, the parameters of this correction stage are frozen during classification training, enforcing a strict separation between bias normalization and cancer decision-making and preventing the classifier from exploiting spurious intensity cues.

In the second stage, the corrected biomarkers are confined to the prostate region, augmented, and decomposed into overlapping $11 \times 11$ patches. These patches are upsampled through a learnable convolutional projection and processed by a ResNet-18 classifier~\cite{he2016deep} to distinguish cancerous from non-cancerous tissue. This decoupled formulation improves robustness to acquisition-dependent artifacts and lesion boundary uncertainty, leading to consistent performance gains over radiomics and end-to-end CNN baselines.

The remainder of this paper is organized as follows. Section~\ref{sec:Background} reviews the background on HM-MRI, PIA-based biomarker generation, and bias field correction. Section~\ref{sec:method} describes the proposed two-stage prostate cancer (PCa) detection framework, and Section~\ref{sec:Results} presents the experimental results and analysis.

\section{Background}
\label{sec:Background}
\subsection{Hybrid Multi-dimensional MRI}
\label{subsec:HMMRI}

Hybrid Multi-Dimensional MRI (HM-MRI) \cite{chatterjee2025hybrid,chatterjee2018diagnosis} is an advanced quantitative imaging technique that integrates two fundamental contrast mechanisms used in multiparametric MRI (mpMRI): $T_2$ relaxometry and diffusion-weighted imaging. By modeling the unique MRI properties of prostate tissue compartments—stroma, epithelium, and lumen-HM-MRI enables noninvasive estimation of tissue composition and microstructure. These quantitative measurements serve as tissue-specific biomarkers that can distinguish malignant from benign regions, providing a deeper physiological interpretation beyond conventional mpMRI signal intensities.

Clinical investigations and reader studies have shown that HM-MRI can enhance radiologists’ diagnostic accuracy and improve MRI-ultrasound fusion biopsy guidance compared to standard PI-RADS-based evaluation \cite{chatterjee2025prospective}. Its standardized acquisition, reproducible analysis, and automated interpretation capabilities make HM-MRI a promising imaging platform for objective, quantitative diagnosis of prostate cancer (PCa), with the potential to reduce unnecessary biopsies and improve clinical decision-making.

\subsection{Physics-Informed Autoencoder} 
\label{subsec:PIA}
Gundogdu et al. introduced the Physics-Informed Autoencoder (PIA) to estimate tissue-based biomarkers for PCa using HM-MRI \cite{gundogdu2025physics}, thereby enhancing radiologists’ diagnostic accuracy. The PIA is a self-supervised deep learning framework designed to estimate prostate tissue microstructure from Hybrid Multi-Dimensional MRI (HM-MRI). Unlike purely data-driven models, PIA integrates a three-compartment diffusion–relaxation model that captures the distinct biophysical properties of the prostate’s epithelial, stromal, and luminal components. By embedding this physical model directly into a neural network, PIA can accurately infer tissue-specific biomarkers—such as volume fractions and relaxation parameters—without requiring large annotated datasets. 

In our framework, we use six of the PIA outputs \cite{gundogdu2025physics}. They are epithelial volume fraction ($v_{ep}$), luminal water fraction ($v_{lu}$), epithelial diffusivity ($d_{ep}$), stromal diffusivity ($d_{st}$), epithelial T2 relaxation time ($t2_{ep}$), and stromal T2 relaxation time ($t2_{st}$). In neural networks, they can be treated as channels for CNN feature extraction, as shown at the input of Fig.~\ref{fig:Model}.

\begin{figure}[t]
    \centering
    \includegraphics[width= 1\linewidth, height= 0.8\linewidth]{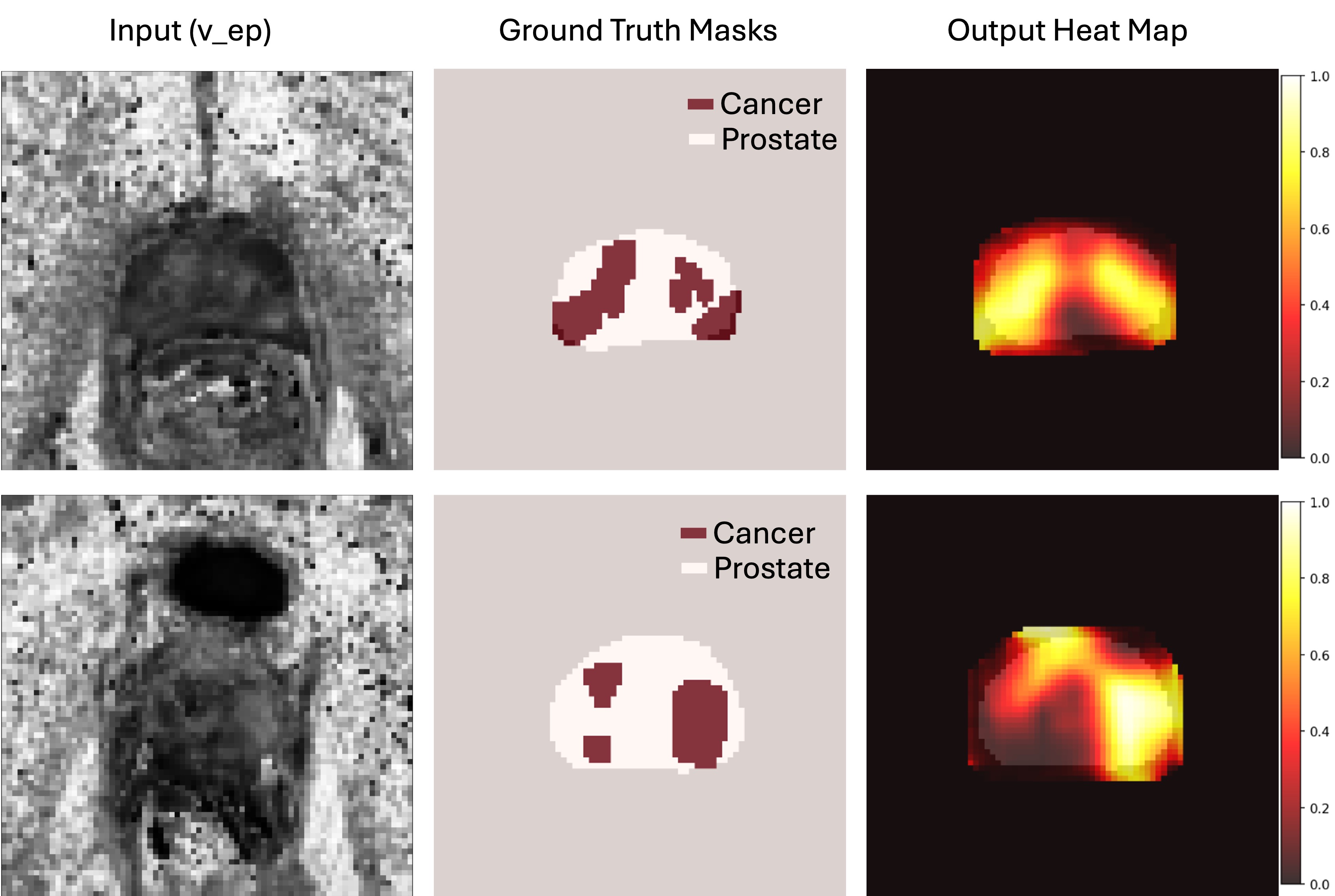}
    \caption{From left to right: epithelial volume fraction ($v_{ep}$) input map, prostate and cancer masks provided by expert radiologists, and the corresponding model-predicted cancer probability heat map. Higher-intensity values indicate regions with a higher likelihood of cancerous tissue.}
\label{fig:Comparison}
\end{figure}

\vspace{-12pt}
\subsection{Hadamard Transform} 
\label{subsec:HT}
The Walsh–Hadamard Transform (WHT) is an orthogonal linear transform that decomposes signals into a basis of binary-valued functions. It has recently been adopted in neural network architectures for efficient feature encoding and bias field correction in MRI \cite{zhu2024probabilistic} to suppress intensity inhomogeneities and enhance tissue contrast normalization. The Hadamard matrix of order $2^m$ is defined recursively as
\begin{equation}
H_m = \frac{1}{\sqrt{2}}
\begin{bmatrix}
H_{m-1} & H_{m-1} \\
H_{m-1} & -H_{m-1}
\end{bmatrix}, \quad H_0 = [1],
\end{equation}
Unlike conventional convolutional filters, Hadamard-based layers exploit transform-domain operations to replace multiplications with sign-based additions, enabling faster and more energy-efficient computation as illustrated in stage 1 of Fig.~\ref{fig:Model}.

Given a tissue biomarker map $X \in \mathbb{R}^{H \times W}$, we model the observed intensity as
\begin{equation}
X = B \odot X^{\text{true}} + \epsilon,
\end{equation}
where $B$ denotes a smooth low-frequency bias field, $\odot$ is element-wise multiplication, and $\epsilon$ represents noise.

Due to its energy compaction property, the Walsh–Hadamard Transform concentrates low-frequency components of $B$ into a small subset of coefficients, enabling efficient suppression of bias-related variations through transform-domain learning.

\section{Methodology}
\label{sec:method}

In this work, we propose using tissue composition maps derived from HM-MRI within a two-stage AI framework for robust and automated Prostate Cancer (PCa) detection: Hadamard-Bias Network + RensNet18 (\textbf{HBR-Net-18}).

\subsection{Stage 1: Bias-Aware Biomarker Correction}

Let $\mathbf{X} = \{X_k\}_{k=1}^{6}$ denote the six tissue-based biomarker maps generated from HM-MRI by the Physics-Informed Autoencoder (PIA). The first stage aims to suppress low-frequency intensity inhomogeneities present in each biomarker map. A probabilistic Hadamard U-Net $f_{\theta}$ is trained to estimate bias-corrected biomarker maps
\begin{equation}
\hat{\mathbf{X}} = f_{\theta}(\mathbf{X}),
\end{equation}
where $\hat{\mathbf{X}} = \{\hat{X}_k\}_{k=1}^{6}$.

Following the approach of Zhu et al.~\cite{zhu2024probabilistic}, reference bias-corrected maps $X_k^{\text{N4}}$ are generated using the N4 bias field correction algorithm (N4ITK)~\cite{5445030}. The network is trained by minimizing
\begin{equation}
\mathcal{L}_{\text{bias}} = \sum_{k=1}^{6} \| \hat{X}_k - X_k^{\text{N4}} \|_2^2 .
\end{equation}

After convergence, the parameters $\theta$ are frozen and the network is used as a deterministic preprocessing module for the downstream classification stage. This enforces a strict separation between bias normalization and cancer detection, preventing the classifier from exploiting acquisition-dependent intensity variations. The bias correction process is illustrated in Fig.~\ref{fig:Model}.

\subsection{Stage 2: Patch-Level Cancer Classification}

From the corrected biomarker maps $\hat{\mathbf{X}}$, prostate-specific regions are first isolated using radiologist-provided masks and expanded by two voxels in all directions to ensure full gland coverage. The prostate volume is then divided into overlapping $11 \times 11$ patches with a stride of 2 across all six biomarker channels, yielding per-slice patch tensors of size $6 \times 11 \times 11$.

To incorporate inter-slice spatial context, each target slice $N$ is augmented with its adjacent slices $(N-1, N, N+1)$. The corresponding patches are stacked along the slice dimension, forming
\begin{equation}
\mathbf{P}_i \in \mathbb{R}^{3 \times 6 \times 11 \times 11}.
\end{equation}

Each patch tensor is passed through a learnable convolutional upsampling module consisting of (i) a projection to $18 \times 16 \times 16$ and (ii) a $3 \times 3$ convolution with batch normalization and ReLU activation to produce feature maps of size $64 \times 32 \times 32$. This learned upsampling preserves fine spatial structure while adapting the patches to the classification network's input dimensionality.

Patch-level cancer probability is then estimated using a ResNet-18 classifier $g_{\phi}$:
\begin{equation}
p_i = g_{\phi}(\mathbf{P}_i), \quad p_i \in [0,1].
\end{equation}

The classifier is trained end-to-end using binary cross-entropy loss
\begin{equation}
\mathcal{L}_{\text{cls}} = -y_i \log(p_i) - (1-y_i)\log(1-p_i),
\end{equation}
where $y_i \in \{0,1\}$ denotes the ground-truth cancer label.

The complete bias correction, patch extraction, upsampling, and classification pipeline is shown in Fig.~\ref{fig:Model}. A qualitative example of an epithelial volume fraction ($v_{ep}$) input, radiologist annotations, and predicted cancer probability heatmap is presented in Fig.~\ref{fig:Comparison}.

The overall framework optimizes cancer detection in a decoupled manner:
\begin{equation}
\min_{\phi} \mathcal{L}_{\text{cls}}(g_{\phi}(f_{\theta}(\mathbf{X}))), \quad \text{with } \theta \text{ fixed}.
\end{equation}

\subsection{Training Strategy}

Both stages were implemented in PyTorch and trained separately on an NVIDIA GPU A6000 workstation. For Stage 1, the Hadamard U-Net was optimized using the AdamW optimizer with a learning rate of $1\times10^{-3}$ for 100 epochs, while Stage 2 employed a learning rate of $1\times10^{-4}$ for the ResNet-18 classifier. Mean-squared error loss is used for intensity correction (stage 1). Focal loss \cite{8417976}, which is used for data imbalance between non-cancer patches and cancer patches, for cancer classification (stage 2). Training and validation were conducted on patient-wise splits to prevent data leakage between subjects.

\section{Experimental Results}
\label{sec:Results}

\subsection{Data Collection and Pre-Processing}

The HM-MRI data used in these experiments was collected from 86 patients who underwent MRI on a 3T Philips MRI scanner and have the outputs of the PIA model. 34 patients are cancer-free, and 52 patients have cancer. The number of slices for each patient ranges from 18 to 30 slices. As the model expects patches for input and output, we use a patch cancer threshold that labels cancer patches as cancer if at least 70\% of the voxels within the patch are actual cancer. Non-cancer patches are extracted from cancer-free patients only. 

Furthermore, after stage 1, the input images are transformed using different augmentation methods to increase the dataset's sample size and enhance image quality. We introduce a horizontal flip with a probability of 50\% and a vertical flip with a probability of 50\%. We additionally use 90-degree rotation with a probability of 50\%. Finally, we introduce three different histogram equalizations. The histogram equalization procedure applied to the epithelial biomarker ($v_{\text{ep}}$) is defined as
\begin{equation}
x'_{v_{\text{ep}}} =
\begin{cases}
0.1\,x_{v_{\text{ep}}}, & x_{v_{\text{ep}}} < T_{\mathrm{d}},\\[4pt]
1, & x_{v_{\text{ep}}} \ge T_{\mathrm{u}},\\[6pt]
\displaystyle \frac{x_{v_{\text{ep}}} - T_{\mathrm{d}}}{T_{\mathrm{u}} - T_{\mathrm{d}}}, & \text{otherwise.}
\end{cases}
\tag{2} \quad ,
\end{equation}
where $T_{\mathrm{d}}$ and $T_{\mathrm{u}}$ denote the lower and upper intensity thresholds, respectively, which were empirically selected from the ranges 
$T_{\mathrm{d}} \in \{0.10, 0.15, 0.20\}$ and $T_{\mathrm{u}} \in \{0.80, 0.85, 0.90\}$ during training and validation. 
The same procedure was applied inversely to the luminal biomarker ($v_{\text{lu}}$) to highlight complementary intensity distributions. 
This adjustment expands the dynamic range and improves contrast between epithelial and luminal tissues prior to model training.

\subsection{Evaluation Results}

\begin{table}[t]
\centering
\caption{\textbf{Patch-Level} prostate cancer detection performance comparison (percentage) obtained after 5 folds.}
\label{tab:model_comparison_patch}
\renewcommand{\arraystretch}{1.2}
\setlength{\tabcolsep}{6pt}
\begin{tabular}{lccc}
\toprule
\textbf{Model} & \textbf{Sensitivity} & \textbf{Specificity} & \textbf{Accuracy} \\
\midrule
Radiomics \cite{campos2025radiomics}           & 91.0& 75.0 & 80.0  \\
ResNet-18~\cite{kang2024deep}   & \textbf{97.9} & 60.5 & 62.5 \\
ResNet-50~\cite{talaat2024improved}   & 91.1 & 87.8 & 82.1 \\
HBR-Net-50 & 92.2 & 90.6 & 91.8  \\
\textbf{HBR-Net-18} & 94.4 & \textbf{93.3} & \textbf{93.8}  \\
\bottomrule
\end{tabular}
\end{table}

\begin{table}[t]
\centering
\caption{\textbf{Patient-Level} prostate cancer detection performance comparison (percentage) obtained after 5 folds.}
\label{tab:model_comparison_patient}
\renewcommand{\arraystretch}{1.2}
\setlength{\tabcolsep}{6pt}
\begin{tabular}{lcccc}
\toprule
\textbf{Model} & \textbf{Sensitivity} & \textbf{Specificity}  & \textbf{Accuracy}  \\
\midrule
Radiomics~\cite{campos2025radiomics}           & \textbf{100.0} & 20.0 & 60.0 \\
ResNet-18~\cite{kang2024deep}   & 71.2 & 71.4 & 72.1 \\
ResNet-50~\cite{talaat2024improved}   & 67.3 & 64.7 & 66.3  \\
HBR-Net-50 & 84.6 & 76.5  & 81.4  \\
\textbf{HBR-Net-18} & 88.5 & \textbf{78.5}  & \textbf{83.7}  \\
\bottomrule
\end{tabular}
\end{table}

We evaluated our results based on sensitivity, specificity, voxel-level overall accuracy, and patient-level cancer and non-cancer accuracy, defined as follows:
$
\mathrm{Sensitivity} = \frac{TP}{TP + FN},
\mathrm{Specificity} = \frac{TN}{TN + FP},
$
and
$
\mathrm{Voxel\text{-}level\ Accuracy} = 
\frac{TP + TN}{TP + TN + FP + FN},
$.
The $
\mathrm{Patient\text{-}level\ Accuracy} = \\
N_{\mathrm{correct\ patients}}/N_{\mathrm{total\ patients,}}$
where $TP$, $TN$, $FP$, and $FN$ denote the number of true positives, true negatives, 
false positives, and false negatives, respectively, and 
$N_{\mathrm{correct\ patients}}$ represents the number of patients 
whose overall cancer/non-cancer label was correctly predicted.

To evaluate the model’s robustness and generalization, we employed a 5-fold cross-validation strategy \cite{yadav2016analysis}. As shown in Table~\ref{tab:model_comparison_patch}, we compare the proposed \textbf{HBR-Net-18}, which utilizes ResNet18, with Radiomics~\cite{campos2025radiomics}, ResNet18~\cite{kang2024deep}, ResNet50~\cite{talaat2024improved}, and HBR-Net-50, which utilizes ResNet50. For fair comparison,  ResNet18 and ResNet50 were trained with raw, un-enhanced data using Stage 2 only. Our \textbf{HBR-Net-18} model improved patch-level accuracy by 17.25\% compared to Radimoics, 50.08\% compared with ResNet18, 14.25\% compared to ResNet50, and 2.18\% compared with HBR-Net-50 while maintaining a high and balanced patch-level sensitivity and specificity. 

Similarly, in Table~\ref{tab:model_comparison_patient}, HB-Net-18 improves performance, in terms of accuracy, by an average of 21\%  across all baseline models. Here, our model outperforms the radiomics results by 292.50\% in terms of specificity towards cancer-free patients while maintaining a high sensitivity towards cancer patients of 88.5\%. 

\subsection{Ablation Study}

To assess the contribution of individual components, we conducted a set of targeted ablation experiments, shown in Table~\ref{tab:ablation}. Removing the bias correction stage resulted in a noticeable performance degradation, highlighting the importance of suppressing low-frequency intensity inhomogeneities prior to classification. Excluding adjacent slices and operating in a purely 2D setting led to reduced performance, indicating that limited 3D contextual information improves spatial consistency in cancer detection. We also evaluated alternative patch sizes and observed that the selected $11\times11$ configuration provides a favorable trade-off between local detail and contextual coverage. These results validate the key design choices of the proposed framework.

\section{Conclusion}

We presented HB-Net-18, a two-stage AI framework for automated prostate cancer detection using HM-MRI and PIA-derived biomarkers. The first stage employed a probabilistic Hadamard U-Net to effectively correct bias field inhomogeneities, improving data consistency. The second stage utilized a ResNet-18 classifier on overlapping patches, integrating 3D spatial information for robust detection. Experimental results demonstrate that the proposed method significantly outperforms standard radiomics and conventional CNN approaches, achieving high and balanced accuracy. However, the current study is limited by a relatively small dataset, which prevents the use of more advanced architectures such as Transformers or Hybrid methods. Further, the current sampling strategy that excluded benign tissue from cancerous patients during training will be revised for a more generic purpose of prostate cancer analysis. Future work will focus on incorporating these benign mimics (e.g., BPH, inflammation) into the training data to improve localization accuracy and enhance clinical robustness.

\begin{table}[t]
\centering
\caption{\textbf{Patch-Level Ablation Study} evaluating the contribution of key components in the proposed HBR-Net-18 framework.}
\label{tab:ablation}
\begin{tabular}{lccc}
\hline
\textbf{Configuration}  & \textbf{Slice Context} & \textbf{Patch Size} & \textbf{Acc} \\
\hline
No Stage-1               & $N\!\pm\!1$ & $11\times11$ & 89.1 \\
2D Only                 & $N$          & $11\times11$ & 88.2 \\
Full HBR-Net-18         & $N\!\pm\!1$ & $9\times9$ & 91.2 \\
Full HBR-Net-18          & $N\!\pm\!1$ & $15\times15$ & 90.9 \\
Full HBR-Net-18         & $N\!\pm\!1$ & $11\times11$ & \textbf{93.8} \\
\hline
\end{tabular}
\end{table}
\section{Compliance with ethical standards}
\label{sec:ethics}
This retrospective study of prospectively collected data was
compliant with the Health Insurance Portability and Ac-
countability Act and conducted after University of Chicago
institutional review board approval and  obtaining
patients’ written informed consent.

\bibliographystyle{IEEEbib}
\bibliography{strings,refs}

\end{document}